# Full-Pose Tracking via Robust Control for Over-Actuated Multirotors

Mohamad Hachem[1], Clément Roos[2], Thierry Miquel[1], and Murat Bronz[1]

*Abstract*— This paper presents a robust cascaded control architecture for over-actuated multirotors. It extends the Incremental Nonlinear Dynamic Inversion (INDI) control combined with structured $\mathcal{H}_\infty$ control, initially proposed for under-actuated multirotors in [12], to a broader range of multirotor configurations. To achieve precise and robust attitude and position tracking, we employ a weighted least-squares geometric guidance control allocation method, formulated as a quadratic optimization problem, enabling full-pose tracking. The proposed approach effectively addresses key challenges, such as preventing infeasible pose references and enhancing robustness against disturbances, as well as considering multirotor's actual physical limitations. Numerical simulations with an over-actuated hexacopter validate the method's effectiveness, demonstrating its adaptability to diverse mission scenarios and its potential for real-world aerial applications.

## I. Introduction

Multirotor aerial robots have gained a lot of attention in recent years. Due to their adaptability and ease of use, they can be applied in a wide range of domains, including human-machine interaction, inspection, photography, payload transportation, and manipulations [1], [2]. In addition, their ability to access hard-to-reach areas makes them invaluable for disaster assessment and environmental monitoring. Each task comes with its own challenges that need to be addressed to ensure safe and successful flights. Constant technological progress is therefore being made to improve their performance, robustness, and potential industrial applications.

Aerial vehicle configurations vary depending on system requirements and performance needs. In particular, reliability and maneuverability are strongly influenced by the actuation set used. In multirotor systems, the geometric distribution of the actuators plays a key role in determining their classification [3]. Systems with distributions that exert forces in only one direction are classified as under-actuated (UA), as not all their degrees of freedom (DoF) can be controlled independently. On the other hand, systems are considered fully-actuated (FA) when all of their DoF can be controlled independently, thanks to an actuator distribution capable of exerting forces and torques along the 6-DoF independently. Finally, over-actuated (OA) systems share the same independent control capabilities as FA systems, but are equipped with redundant actuators, making them more robust and reliable against actuator failures and saturations. Within this classification, each category is associated with its own control approaches and challenges, ranging from coupling rotational and translational dynamics to achieving fully decoupled dynamics control.

FA and OA multirotors are classified as Laterally Bounded Input Forces (LBIF) systems [4], for which saturations in the lateral dynamics can be easily reached depending on the position reference. A full-pose geometric controller on $SE(3)$ is proposed in [4] using feedback linearization techniques, also known as Nonlinear Dynamics Inversion (NDI), to track predefined position and orientation trajectories. The lateral forces are assumed to be bounded by a geometric constraint contained within the attainable force set (AFS). Building on this concept, an optimization-based control method is proposed by [5] to achieve full-pose trajectory tracking by searching for a solution directly in the entire AFS. In both methods, a feasible position and attitude trajectory is recomputed to account for physical limitations. Among other contributions, [6] combines explicit reference governors with NDI to address actuator saturation and enhance system stability. In addition, NDI with dynamic extension for OA quadcopters featuring rotated motors is applied by [7], without encountering a full-pose tracking problem. The same controller is used by [8] for a FA hexacopter with constant motor inclinations, and an optimal analysis of the motor orientation is proposed. A detailed experimental comparison is presented in [9] between NDI, quaternion-based P/PID controllers, and PI stabilizers with feedforward controllers for position tracking, highlighting the strengths of each approach. It emphasizes the importance of a control strategy that accounts for the unavoidable cross-coupling between position and attitude dynamics, which might affect performance during aggressive maneuvers. Finally, a cascaded $\mathcal{H}_\infty$ control structure for FA multirotors is proposed in [10]. The design assumes fully coupled rotational and translational dynamics, using distinct $\mathcal{H}_\infty$ weights to balance acceleration distribution between attitudes and forces, while minimizing rotations.

NDI is one of the most commonly used control methods, especially in the context of FA and OA multirotors.

However, it requires an accurate model for inversion, and its robustness is tied to the linear controller managing the linear regime. Moreover, applications involving human interaction or operations in confined environments require effective disturbance rejection to ensure safety and reliability. Improving robustness to disturbances is therefore crucial for mission success. In this perspective, robust controllers have been extensively studied for UA multirotors, with promising experimental results. In particular, Incremental Nonlinear Dynamic Inversion (INDI) has shown good robustness against disturbances and modeling uncertainties [11], especially when combined with reduced-order $\mathcal{H}_\infty$ controllers [12]. The latter paper shows that INDI/$\mathcal{H}_\infty$ controllers improve system stability and robustness to disturbances, while removing the need for an accurate knowledge of multirotor's mass and inertia, as the control laws rely only on the actuator model. Furthermore, the INDI/$\mathcal{H}_\infty$ control ability to operate at high frequencies offers a significant computational advantage over optimization-based methods like NMPC and robust MPC, making it highly suitable for real-time embedded control applications. **In this work, we generalize the INDI/$\mathcal{H}_\infty$ control architecture of [12], initially proposed for UA multirotors, to accommodate various multirotor configurations including the FA and OA classes. Our main contribution is to address the significant lack of robust architectures for OA and FA multirotors, which is crucial for ensuring safety in modern applications. Additionally, we employ geometric weighted least squares (WLS) control allocation to achieve full-pose tracking with the ability to run at high frequencies (>500 Hz). The proposed approach considers a varying subset of the AFS to achieve feasible forces and prevent the tracking of infeasible poses , while prioritizing position over attitude tracking, particularly in the presence of disturbances (e.g. gusts, downwash).**

The rest of the paper is organized as follows. Section II provides the mathematical modeling of a general OA multirotor system. Section III introduces the proposed INDI/$\mathcal{H}_\infty$ architecture with a focus on control allocation. Numerical validation results are presented and discussed in Section IV for an OA hexacopter, and Section V outlines the main conclusions and future research directions.

## II. Modeling
### A. Multirotor model

We consider a 6-DoF rigid multirotor system with mass $m$ and inertia matrix $\boldsymbol{I}_B$. The primary reference frames are the world (inertial) frame $\mathcal{F}_W$, based on the North-East-Down convention with basis vectors $[\boldsymbol{x}_W, \boldsymbol{y}_W, \boldsymbol{z}_W]$, the body frame $\mathcal{F}_B$, aligned with the multirotor's body axes $[\boldsymbol{x}_B, \boldsymbol{y}_B, \boldsymbol{z}_B]$, and the motor frame $\mathcal{F}_{M_i}$ associated with the $i^{th}$ motor, with axes $[\boldsymbol{x}_{M_i}, \boldsymbol{y}_{M_i}, \boldsymbol{z}_{M_i}]$. The rotation from $\mathcal{F}_B$ to $\mathcal{F}_W$ can be described using the Euler angles $\boldsymbol{\mu} = [\phi, \theta, \psi]^T$ through the rotation matrix $\boldsymbol{R}_B^W \in \mathcal{SO}(3)$ with $(XYZ)$ convention. Similarly, the rotation from $\mathcal{F}_{M_i}$ to $\mathcal{F}_B$ can be parameterized by $\boldsymbol{\lambda}_i = [\alpha_i, \beta_i, \gamma_i]^T$ through the rotation matrix $\boldsymbol{R}_{M_i}^B \in \mathcal{SO}(3)$, as illustrated in Fig. 1. The multirotor's position, velocity and acceleration in $\mathcal{F}_W$ are denoted by $\boldsymbol{\xi} = [x, y, z]^T$, $\dot{\boldsymbol{\xi}} = \boldsymbol{v} = [v_x, v_y, v_z]^T$ and $\ddot{\boldsymbol{\xi}} = \boldsymbol{a} = [a_x, a_y, a_z]^T$ respectively. Its angular speed and acceleration in $\mathcal{F}_B$ are denoted by $\boldsymbol{\Omega} = [p, q, r]^T$ and $\dot{\boldsymbol{\Omega}} = [\dot{p}, \dot{q}, \dot{r}]^T$. Applying Newton-Euler equations of motion leads to the dynamics:

$$\ddot{\boldsymbol{\xi}} = g\boldsymbol{z}_W + \frac{1}{m}\bigl(\boldsymbol{R}_B^W \boldsymbol{f}_b + \boldsymbol{f}_d\bigr) \quad (1)$$

$$\dot{\boldsymbol{R}}_B^W = \boldsymbol{R}_B^W S(\boldsymbol{\Omega}) \, , \, \dot{\boldsymbol{\Omega}} = \boldsymbol{I}_B^{-1}\bigl(\boldsymbol{\tau}_b - \boldsymbol{\Omega} \times \boldsymbol{I}_B \boldsymbol{\Omega} + \boldsymbol{\tau}_d\bigr) \quad (2)$$

where $S(.)$ is the skew-symmetric operator, $\boldsymbol{f}_d$ and $\boldsymbol{\tau}_d$ denote external perturbation forces and torques, and $g$ is the constant gravitational acceleration. The force generated by the $i^{th}$ motor is directed along $\boldsymbol{z}_{M_i}$ and is expressed as $f_{m_i} = k_t \omega_i^2$, where $k_t$ is the lift coefficient and $\omega_i$ is the $i^{th}$ motor angular speed. Its components in $\mathcal{F}_B$ are given by $\boldsymbol{f}_{b_i} = \boldsymbol{R}_{M_i}^B f_{m_i} \boldsymbol{z}_{M_i}$. The torque $\tau_{m_i}$ around $\boldsymbol{z}_{M_i}$ is expressed as $\tau_{m_i} = c_i k_d \omega_i^2$ where $k_d$ is the torque coefficient, and $c_i = \pm 1$ corresponds to the rotational direction of the $i^{th}$ motor. The resulting total forces $\boldsymbol{f}_b$ and torques $\boldsymbol{\tau}_b$ in $\mathcal{F}_B$ are then calculated as follows:

$$\begin{bmatrix} \boldsymbol{f}_b \\ \boldsymbol{\tau}_b \end{bmatrix} = \underbrace{\begin{bmatrix} \boldsymbol{I}_3 & \dots & \boldsymbol{I}_3 \\ S(\boldsymbol{d}_1) + rc_1\boldsymbol{I}_3 & \dots & S(\boldsymbol{d}_n) + rc_n\boldsymbol{I}_3 \end{bmatrix}}_{\boldsymbol{B}} \boldsymbol{u} \quad (3)$$

where $\boldsymbol{u} = \begin{bmatrix} \boldsymbol{f}_{b_1}^T & \dots & \boldsymbol{f}_{b_n}^T \end{bmatrix}^T$, $n$ is the number of motors, $\boldsymbol{d}_i \in \mathbb{R}^3$ is the position of the $i^{th}$ motor in $\mathcal{F}_B$, and $r = k_d/k_t$. The allocation matrix $\boldsymbol{B} = \begin{bmatrix} \boldsymbol{B}_1^T & \boldsymbol{B}_2^T \end{bmatrix}^T \in \mathbb{R}^{6 \times 3n}$ in Eq. (3) therefore maps $\boldsymbol{f}_b$ and $\boldsymbol{\tau}_b$ to the forces exerted by the motors. It is composed of the force and torque allocation matrices $\boldsymbol{B}_1 \in \mathbb{R}^{3 \times 3n}$ and $\boldsymbol{B}_2 \in \mathbb{R}^{3 \times 3n}$:

$$\begin{bmatrix} \boldsymbol{f}_b \\ \boldsymbol{\tau}_b \end{bmatrix} = \boldsymbol{B}\boldsymbol{u} \text{ where } \boldsymbol{f}_b = \boldsymbol{B}_1 \boldsymbol{u} \text{ and } \boldsymbol{\tau}_b = \boldsymbol{B}_2 \boldsymbol{u} \quad (4)$$

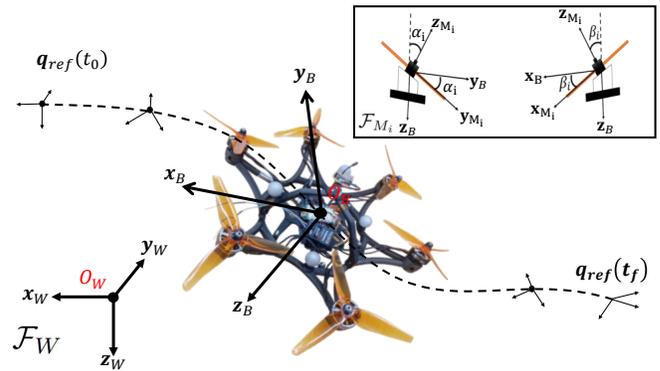

Fig. 1: Illustration of the multirotor and main frames.

### B. Attainable and Feasible Force Sets

The AFS is the set of all forces $\boldsymbol{f}_b = \boldsymbol{B}_1 \boldsymbol{u}$ the multirotor can generate when $\boldsymbol{u} \in \mathcal{U}$, the set of admissible inputs, obtained by considering the physical limitations of the motor angular speeds $\omega_i$ and rotations $\boldsymbol{\lambda}_i$. The set of feasible forces at a specific torque $\boldsymbol{\tau}_{b_s}$ is a subset of the AFS defined as in [5]:

$$\mathcal{S}(\boldsymbol{\tau}_{b_s}) = \left\{ \boldsymbol{f}_b \in \mathbb{R}^3 : \begin{bmatrix} \boldsymbol{f}_b \\ \boldsymbol{\tau}_{b_s} \end{bmatrix} = \begin{bmatrix} \boldsymbol{B}_1 \\ \boldsymbol{B}_2 \end{bmatrix} \boldsymbol{u}, \boldsymbol{u} \in \mathcal{U} \right\} \quad (5)$$

In particular, $\mathcal{S}(\boldsymbol{\tau}_{b_s} = 0)$ corresponds to the set of forces that can be achieved without performing rotations:

$$\mathcal{S}(\boldsymbol{\tau}_{b_s} = 0) = \{ \boldsymbol{f}_b \in \mathbb{R}^3 : \boldsymbol{f}_b = \boldsymbol{B}_1 \boldsymbol{u}, \boldsymbol{u} \in \mathcal{U} \cap \ker(\boldsymbol{B}_2) \} \quad (6)$$

This set defines the attainable forces within the null space of $B_2$, characterizing multirotor capabilities in hovering configurations. While setting $\boldsymbol{\tau}_{b_s} = 0$ may limit maneuverability in some applications (specifically for FA multirotors) compared to [5], it ensures the guidance controller generates feasible solutions for stabilization (see Section III-C). Moreover, in contrast to [4], where a constant geometric shape (e.g. a cylinder) is assumed for the AFS subset, our approach directly searches within the null space of $\boldsymbol{B}_2$.

### C. Full-Pose Tracking

The full-pose tracking problem is formulated as in [5]. A full-pose reference is denoted by $\boldsymbol{q}_{ref}(t) = (\boldsymbol{\xi}_{ref}(t), \boldsymbol{R}_{ref_B}^W(t))$, where $\boldsymbol{\xi}_{ref}(t) \in \mathbb{R}^3$ represents the reference position, assumed to be smooth up to order 4 to ensure a continuous and differentiable jerk, and $\boldsymbol{R}_{ref_B}^W(t)$ is the reference rotation. For the remainder of the paper, we will omit the explicit time dependence ($t$) for clarity. A pose is considered feasible if:

$$\boldsymbol{f}_{b_{ref}} = \boldsymbol{R}_{ref_B}^{W^T} \boldsymbol{f}_{w_{ref}} \in \mathcal{S}(\boldsymbol{\tau}_{b_{ref}}) \quad (7)$$

where $\boldsymbol{f}_{w_{ref}}$ is the force in $\mathcal{F}_W$ required to follow $\boldsymbol{\xi}_{ref}$, $\boldsymbol{f}_{b_{ref}}$ its equivalent in $\mathcal{F}_B$, and $\boldsymbol{\tau}_{b_{ref}}$ is the torque required to follow $\boldsymbol{R}_{ref_B}^W$. If not, a commanded feasible pose $\boldsymbol{q}_c = (\boldsymbol{\xi}_c, \boldsymbol{R}_{c_B}^W)$ is used instead, which satisfies:

$$\boldsymbol{\xi}_c = \boldsymbol{\xi}_{ref}$$
$$\boldsymbol{R}_{c_B}^W = \arg\min \ \text{dist}(\boldsymbol{R}_B^W, \boldsymbol{R}_{ref_B}^W) \quad (8)$$
$$\text{subject to} \quad \boldsymbol{f}_{b_c} = \boldsymbol{R}_{c_B}^{W^T} \boldsymbol{f}_{w_c} \in \mathcal{S}(\boldsymbol{\tau}_{b_c})$$

where $\text{dist}(\boldsymbol{R}_B^W, \boldsymbol{R}_{ref_B}^W)$ is a distance function to be determined, see Section III-C. The objective is therefore to follow the reference position, while following a feasible orientation as close as possible to the reference one. Throughout the rest of the paper, rotations are described using Euler angles without loss of generality for the proposed problem.

### III. VEHICLE CONTROL

The cascaded control architecture used in this work is shown in Fig. 2. It consists of an inner controller that handles rotational dynamics (stabilization loop) and an outer controller that manages translational dynamics (guidance loop). The control allocation is responsible for distributing the incremental virtual commanded control inputs $\delta\boldsymbol{\nu}_c = [\delta f_{b_{x_c}}, \delta f_{b_{y_c}}, \delta f_{b_{z_c}}, \delta\tau_{\phi_c}, \delta\tau_{\theta_c}, \delta\tau_{\psi_c}]^T$ over the actual actuator inputs *i.e.* for computing the angular velocity $\omega_i$ and the orientation $\boldsymbol{\lambda}_i$ of each actuator, as explained in Section III-A. The stabilization controller is responsible for tracking the commanded feasible attitude $\boldsymbol{\mu}_c$ provided by the guidance loop. It consists of an INDI controller to invert the rotational dynamics and a structured $\mathcal{H}_\infty$ controller to enhance robustness to disturbances, as presented in [12]. The guidance loop takes as input the reference position $\boldsymbol{\xi}_{ref}$ and orientation $\boldsymbol{\mu}_{ref}$. It consists of both an INDI and a structured $\mathcal{H}_\infty$ controller described in Section III-B, followed by a WLS control allocation presented in Section III-C to generate $\boldsymbol{\mu}_c$ and $\delta\boldsymbol{f}_{b_c} = [\delta f_{b_{x_c}}, \delta f_{b_{y_c}}, \delta f_{b_{z_c}}]^T$.

### A. Control Allocation

In this section, we present a general allocation matrix formulation for multirotors, assuming that the frame $\mathcal{F}_{M_i}$ of each of the $n$ motors has 3-DoF of rotations ($\boldsymbol{\lambda}_i$) with respect to the body frame $\mathcal{F}_B$. Referring back to Eq. (3), it is possible to redefine an allocation matrix $\boldsymbol{B}'(\boldsymbol{\lambda})$ as:

$$\boldsymbol{B}'(\boldsymbol{\lambda}) = [\boldsymbol{B}'_1(\boldsymbol{\lambda}_1), \ldots, \boldsymbol{B}'_n(\boldsymbol{\lambda}_i)] \in \mathbb{R}^{6 \times 3n} \quad (9)$$

where:

$$\boldsymbol{B}'_i(\boldsymbol{\lambda}_i) = \begin{bmatrix} \boldsymbol{I}_3 \\ S(\boldsymbol{d}_i) + rc_i \boldsymbol{I}_3 \end{bmatrix} \boldsymbol{R}_{M_i}^B \in \mathbb{R}^{6 \times 3} \quad (10)$$

The control allocation problem is then formulated as:

$$\underbrace{\begin{bmatrix} \boldsymbol{f}_b \\ \boldsymbol{\tau}_b \end{bmatrix}}_{\boldsymbol{\nu}} = \boldsymbol{B}'(\boldsymbol{\lambda}) \underbrace{\begin{bmatrix} f_{m_1} \boldsymbol{z}_{M_1} \\ \vdots \\ f_{m_n} \boldsymbol{z}_{M_n} \end{bmatrix}}_{F(\boldsymbol{\omega})} \quad (11)$$

As shown in Fig. 2, the guidance and the stabilization INDI controllers generate $\delta\boldsymbol{\nu}_c$ by linearizing the multirotor's dynamics around the measured states. Therefore, the idea is to use a first-order Taylor's expansion of the nonlinear function $\boldsymbol{B}'(\boldsymbol{\lambda}) F(\boldsymbol{\omega})$ around the measured control actuator inputs. This linearization does not impact the system's performance because INDI controllers are usually implemented at high frequencies, thus the trigonometric nonlinearities in $\boldsymbol{B}'(\boldsymbol{\lambda})$ vary relatively small between two consecutive samples. Linearizing Eq. (11) yields:

$$\boldsymbol{\nu} = \underbrace{\boldsymbol{B}'(\boldsymbol{\lambda}_m) F(\boldsymbol{\omega}_m)}_{\boldsymbol{\nu}_m} + \underbrace{\frac{\partial \boldsymbol{B}'(\boldsymbol{\lambda}) F(\boldsymbol{\omega})}{\partial \boldsymbol{u}_a} \bigg|_{\boldsymbol{u}_a = \boldsymbol{u}_{a_m}}}_{\boldsymbol{B}''(\boldsymbol{u}_{a_m})} (\boldsymbol{u}_a - \boldsymbol{u}_{a_m}) \quad (12)$$

where $\boldsymbol{u}_a = [\boldsymbol{\lambda}_1^T, \omega_1, \ldots, \boldsymbol{\lambda}_n^T, \omega_n]^T \in \mathbb{R}^{4n}$ denotes the motor control inputs, and the subscript $m$ stands for "filtered and measured". A (typically second-order) filter $H(s)$ is indeed needed for noise attenuation, as explained in [11], and it is applied to all measured or estimated signals to maintain synchronization. Eq. (12) can then be expressed as:

$$\delta\boldsymbol{\nu} = \boldsymbol{B}''(\boldsymbol{u}_{a_m})(\boldsymbol{u}_a - \boldsymbol{u}_{a_m}) \quad (13)$$

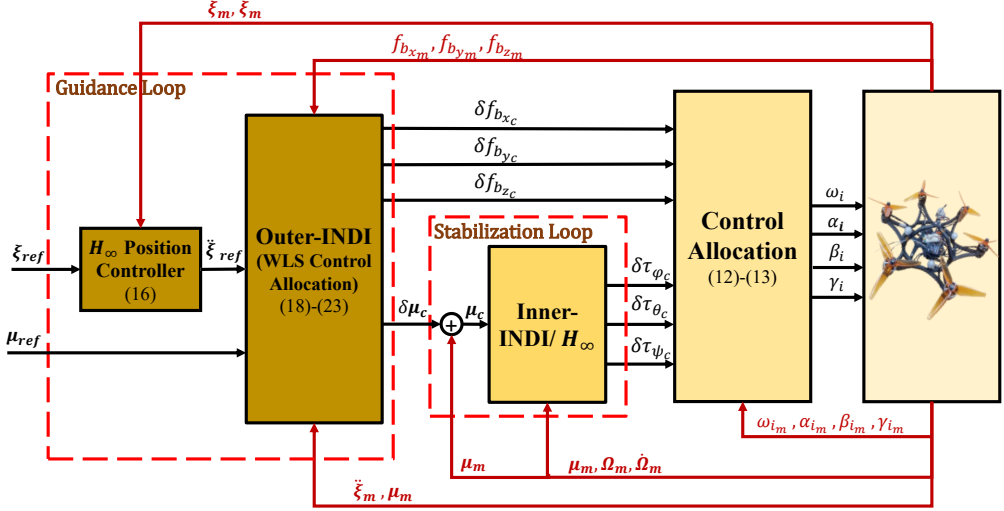

Fig. 2: Robust INDI/$\mathcal{H}_\infty$ cascaded control architecture for full-pose tracking of over-actuated multirotors, where the subscripts $m, ref, c$ denote filtered measured signals, reference signals and commanded values computed by the controllers respectively.

where $\boldsymbol{B}''(\boldsymbol{u}_{a_m}) \in \mathbb{R}^{6\times 4n}$ is the matrix containing the partial derivatives of $\boldsymbol{B}'(\boldsymbol{\lambda})\boldsymbol{F}(\boldsymbol{\omega})$ around the filtered measured control inputs $\boldsymbol{u}_{a_m}$, and $\delta\boldsymbol{\nu} = \boldsymbol{\nu} - \boldsymbol{\nu}_m$. The control inputs $\boldsymbol{u}_a$ to be sent to the motors can finally be computed by inverting $\boldsymbol{B}''(\boldsymbol{u}_{a_m})$. In this work, we use the same control allocation method as for the guidance outer-loop to solve Eq. (13), see Section III-C.

*B. Guidance Controller*

The primary objective of the guidance controller is to ensure accurate tracking of the reference position $\boldsymbol{\xi}_{ref}$ and the reference orientation $\boldsymbol{\mu}_{ref}$, whenever feasible. The input to the translational dynamics model described in Eq. (1) is given by $\boldsymbol{u}_\xi = [f_{b_x}, f_{b_y}, f_{b_z}, \phi, \theta]^T$. The heading angle, $\psi$, is not included in $\boldsymbol{u}_\xi$, as it is usually predefined by a trajectory feedforward controller. After expressing $\boldsymbol{f}_b$ in $\mathcal{F}_W$ as $\boldsymbol{f}_w(\boldsymbol{u}_\xi) = \boldsymbol{R}_B^W \boldsymbol{f}_b$, Eq. (1) is linearized using Taylor's expansion around the filtered input $\boldsymbol{u}_{\xi_m}$, as follows:

$$\ddot{\boldsymbol{\xi}} = g\boldsymbol{z}_W + \frac{1}{m}(\boldsymbol{f}_w(\boldsymbol{u}_{\xi_m}) + \boldsymbol{f}_d) + \frac{1}{m}\frac{\partial \boldsymbol{f}_w(\boldsymbol{u}_\xi)}{\partial \boldsymbol{u}_\xi}\bigg|_{\boldsymbol{u}_\xi = \boldsymbol{u}_{\xi_m}}(\boldsymbol{u}_\xi - \boldsymbol{u}_{\xi_m}) \quad (14)$$

Note that in practice, the forces $\boldsymbol{f}_{b_m} = [f_{b_{x_m}}, f_{b_{y_m}}, f_{b_{z_m}}]^T$ are not directly measured, but can be estimated using Eq. (4) with the knowledge of the filtered actuator states. The first three terms in Eq. (14) represent the filtered acceleration $\ddot{\boldsymbol{\xi}}_m$, while the partial derivatives form the control effectiveness matrix $\boldsymbol{G}(\boldsymbol{u}_\xi)$ of the guidance loop, defined as:

$$\boldsymbol{G}(\boldsymbol{u}_\xi) = \left[\frac{\partial \boldsymbol{f}_w(\boldsymbol{u}_\xi)}{\partial f_{b_x}}, \frac{\partial \boldsymbol{f}_w(\boldsymbol{u}_\xi)}{\partial f_{b_y}}, \frac{\partial \boldsymbol{f}_w(\boldsymbol{u}_\xi)}{\partial f_{b_z}}, \frac{\partial \boldsymbol{f}_w(\boldsymbol{u}_\xi)}{\partial \phi}, \frac{\partial \boldsymbol{f}_w(\boldsymbol{u}_\xi)}{\partial \theta}\right]$$

The outer-INDI control law is obtained by setting $\ddot{\boldsymbol{\xi}} = \ddot{\boldsymbol{\xi}}_{ref}$ in Eq. (14):

$$m\delta\ddot{\boldsymbol{\xi}} = \boldsymbol{G}(\boldsymbol{u}_{\xi_m})\delta\boldsymbol{u}_{\xi_c} \quad (15)$$

Here, $\delta\ddot{\boldsymbol{\xi}} = \ddot{\boldsymbol{\xi}}_{ref} - \ddot{\boldsymbol{\xi}}_m$, $\delta\boldsymbol{u}_{\xi_c}$ is the commanded incremental control input, and $\ddot{\boldsymbol{\xi}}_{ref}$ is computed using a robust linear $\mathcal{H}_\infty$ controller as shown in Fig. 2. This controller is chosen to have a cascaded structure:

$$\ddot{\boldsymbol{\xi}}_{ref} = K_{\dot{\xi}}(s)(K_\xi(s)(\boldsymbol{\xi}_{ref} - \boldsymbol{\xi}_m) - \dot{\boldsymbol{\xi}}_m) \quad (16)$$

to enable both position and velocity tracking. $K_\xi(s)$ and $K_{\dot{\xi}}(s)$ are two first-order systems, designed using the methodology proposed in [12].

The effectiveness matrix $\boldsymbol{G}(\boldsymbol{u}_{\xi_m}) \in \mathbb{R}^{3\times 5}$ in the case of OA and FA multirotors, and its inversion in Eq. (15) leads to an infinite number of possible solutions for $\delta\boldsymbol{u}_{\xi_c}$. Additional constraints can thus be taken into account, leading to a control allocation problem detailed in the next section.

*C. Guidance Control Allocation*

The value of $\delta\boldsymbol{u}_{\xi_c}$ in Eq. (15) can be obtained using a variety of methods, since the problem is formulated as a general control allocation problem as presented by [13]. Methods based on the (weighted) Moore-Penrose pseudo-inverse are the simplest, but they usually do not consider control input saturations [14]. Null-space control allocation is used in the guidance controller for OA vehicles in [15], [16], where the authors assume that the trajectory of the multirotor is predefined to be feasible. Optimization-based methods are also commonly used, as the control allocation problem essentially involves minimizing a defined norm. A detailed comparison of several control allocation methods is presented in [17]. In particular, several least squares techniques are evaluated, using various solvers. Among them, the Weighted Least Squares (WLS) method consists of minimizing a quadratic cost function subject to linear constraints. Proposed by [18], it has been successfully applied on numerous occasions, for example to different unmanned aerial vehicles (UAV) actuation systems by [19], [20].

Eq. (15) is tackled in this context considering the full-pose tracking problem stated in Section II-C. A WLS problem is formulated to ensure the tracking of the commanded trajectory $\boldsymbol{q}_c$, defined as the feasible trajectory closest to $\boldsymbol{q}_{ref}$. The objective is to prioritize the tracking of $\boldsymbol{\xi}_{ref}$ over $\boldsymbol{\mu}_{ref}$ by appropriately selecting the weights in the cost function:

$$C(\delta\boldsymbol{u}_{\xi_c}) = \gamma_{opt} \left\| \boldsymbol{W}_v \left( \boldsymbol{G}(\boldsymbol{u}_{\xi_m})\delta\boldsymbol{u}_{\xi_c} - m\delta\ddot{\boldsymbol{\xi}} \right) \right\|^2 + \left\| \boldsymbol{W}_u \left( \delta\boldsymbol{u}_{\xi_c} - \delta\boldsymbol{u}_{\xi_p} \right) \right\|^2 \quad (17)$$

where $\gamma_{opt}$ is a scalar, and $\boldsymbol{W}_u \in \mathbb{R}^{5\times 5}$ and $\boldsymbol{W}_v \in \mathbb{R}^{3\times 3}$ are two positive definite weighting matrices used to handle the trade-off between the different terms of the cost function. The first term minimizes the error between $m\delta\ddot{\boldsymbol{\xi}}$ and $\boldsymbol{G}(\boldsymbol{u}_{\xi_m})\delta\boldsymbol{u}_{\xi_c}$, thus ensuring accurate position tracking. The second term focuses on tracking a desired control input $\delta\boldsymbol{u}_{\xi_p}$, including the reference orientation $\boldsymbol{\mu}_{ref}$ and given by $\delta\boldsymbol{u}_{\xi_p} = [f_{b_{x_m}}, f_{b_{y_m}}, f_{b_{z_m}}, \phi_{ref} - \phi_m, \theta_{ref} - \theta_m]^T$. To ensure tracking of the heading $\psi_{ref}$, it would be possible to include it in $\delta\boldsymbol{u}_{\xi_c}$ as well via the error $\psi_{ref} - \psi_m$. The function $\text{dist}(\boldsymbol{R}_B^W, \boldsymbol{R}_{ref_B}^W)$, introduced in Section II-C, is appropriately incorporated as $\|\delta\boldsymbol{\mu}_c - (\boldsymbol{\mu}_{ref} - \boldsymbol{\mu}_m)\|$ into the second term of Eq. (17). Additionally, linear constraints are incorporated to limit the lateral forces of the multirotor to ensure having a feasible $\boldsymbol{q}_c$. Minimizing $C(\delta\boldsymbol{u}_{\xi_c})$ under these constraints can then be formulated as a Quadratic Programming (QP) problem, as shown below:

$$\delta\boldsymbol{u}_{\xi_c}^* = \arg\min_{\delta\boldsymbol{u}_{\xi_c}} \left\| \boldsymbol{A}\delta\boldsymbol{u}_{\xi_c} - \boldsymbol{b} \right\|^2 \quad \text{s.t.} \quad \delta\boldsymbol{u}_{\xi_c} \in \mathcal{U}_{\boldsymbol{\xi}_c} \quad (18)$$

where $\delta\boldsymbol{u}_{\xi_c}^*$ represents the optimal solution, $\mathcal{U}_{\boldsymbol{\xi}_c}$ is the admissible feasible set of $\delta\boldsymbol{u}_{\xi_c}$, and $\boldsymbol{A}$ and $\boldsymbol{b}$ are defined as:

$$\boldsymbol{A} = \begin{bmatrix} \gamma_{opt}^{\frac{1}{2}} \boldsymbol{W}_v \boldsymbol{G}(\boldsymbol{u}_{\xi_m}) \\ \boldsymbol{W}_u \end{bmatrix}, \quad \boldsymbol{b} = \begin{bmatrix} \gamma_{opt}^{\frac{1}{2}} \boldsymbol{W}_v \delta\ddot{\boldsymbol{\xi}} \\ \boldsymbol{W}_u \delta\boldsymbol{u}_{\xi_p} \end{bmatrix} \quad (19)$$

The components of the admissible feasible set $\mathcal{U}_{\boldsymbol{\xi}_c}$ are primarily computed based on the set $\mathcal{S}(\boldsymbol{\tau}_{b_s} = 0)$ presented in Section II-B. This is achieved by first defining $\boldsymbol{f}_{w_{ref}}$ as:

$$\boldsymbol{f}_{w_{ref}} = m\delta\ddot{\boldsymbol{\xi}} + \boldsymbol{R}_{m_B}^W \boldsymbol{f}_{b_m} \quad (20)$$

and then computing its relative vector in $\mathcal{F}_B$, considering the current orientation of the multirotor as:

$$\boldsymbol{f}_{b_{ref}} = \boldsymbol{R}_{m_B}^{W^T} \boldsymbol{f}_{w_{ref}} \quad (21)$$

After that, we define a plane $(P)$ parallel to the plane $(f_{b_x} f_{b_y})$ by the equation:

$$(P): f_{b_z} = f_{b_{z_{ref}}} \quad (22)$$

The intersection between the plane $(P)$ and the polyhedron defined by $\mathcal{S}(\boldsymbol{\tau}_{b_s} = 0)$ is denoted as $(P_{xy})$. It is used to constrain the solution of the forces $(f_{b_{x_c}}, f_{b_{y_c}})$, and an example is shown in Fig. 3. The optimization problem in Eq. (18) is then solved with $\mathcal{U}_{\boldsymbol{\xi}_c}$ formed as:

$$A_{xy}(\delta f_{x_c}, \delta f_{y_c}) \leq B_{xy} - A_{xy}(\delta f_{x_m}, \delta f_{y_m}),$$
$$\underline{f_{b_z}} - f_{b_{z_m}} \leq \delta f_{b_{z_c}} \leq \overline{f_{b_z}} - f_{b_{z_m}}, \quad (23)$$
$$\underline{\boldsymbol{\mu}} - \boldsymbol{\mu}_m \leq \delta\boldsymbol{\mu}_c \leq \overline{\boldsymbol{\mu}} - \boldsymbol{\mu}_m.$$

where $A_{xy}, B_{xy}$ represent the boundaries of $(P_{xy})$, and the notation $\underline{\times}$ and $\overline{\times}$ refers to the minimum and maximum allowed values of $\times$, respectively. The solution can be obtained using various solvers, such as Interior Point Optimizer (IPOPT), Trust-Region Reflective Algorithm or Active-Set Method.

## IV. NUMERICAL VALIDATION

To validate the control architecture proposed in Section III, simulations are conducted under different scenarios for an OA hexacopter. They are performed in MATLAB/SIMULINK using a 6-DoF nonlinear simulator at a frequency of 500 Hz running on a PC with an Intel Core i7 12th-generation processor (64-bit). The considered multirotor has $n = 6$ motors, each equipped with two servos, allowing control over the rotation angles $\alpha_i$ and $\beta_i$. In this case, the morphing angle $\gamma_i$ is kept constant, as it primarily influences the attainable torque set and enhances the multirotor's robustness against motor failures [21]. The physical parameters used in the simulations are listed in Table I, where $l$ is the distance from the center of gravity to the motors.

TABLE I: Multirotor physical parameters

| $I_x = 0.0041$ kg.m$^2$ | $I_y = 0.0048$ kg.m$^2$ | $I_z = 0.0599$ kg.m$^2$ |
|---|---|---|
| $m = 0.6656$ kg | 0 rpm $\leq \omega_i \leq 2000$ rpm | $-40° \leq \alpha_i, \beta_i \leq 40°$ |
| $l = 0.15$ m | $k_t = 1.14 \times 10^{-4}$ N/rad/s$^2$ | $k_d = 1.14 \times 10^{-6}$ Nm/rad/s$^2$ |

A first-order actuator model $A(s)$ is used, with the same time constant $\tau_{ac} = \frac{1}{53.94}$ s as in [11]. The second-order filter $H(s)$ introduced in Section III-A has a damping ratio of $\xi_n = 0.55$ and a natural frequency of $\omega_n = 50\,\text{rad/s}$:

$$A(s) = \frac{1}{\tau_{ac}s + 1} \quad \text{and} \quad H(s) = \frac{\omega_n^2}{s^2 + 2\xi_n\omega_n s + \omega_n^2} \quad (24)$$

The control allocation problem of Eq. (18) is solved using MATLAB's `lsqlin` function and the 'Active-Set' solver, with $\gamma_{\text{opt}} = 1000$, $\boldsymbol{W}_u = \text{diag}([1,1,1,10,10,1])$, and $\boldsymbol{W}_v = \text{diag}([100,100,10])$. Here, $\boldsymbol{W}_u \in \mathbb{R}^{6\times 6}$ since it also incorporates the multirotor's heading angle $\psi$. The AFS for the multirotor under consideration is displayed in Fig. 3 and almost coincides with the set $\mathcal{S}(\boldsymbol{\tau}_{b_s} = 0)$ defined in Eq. (6) (this is shown for a particular value of $f_{b_z}$, but the same is observed for all values). This is primarily due to the ability of each motor to independently generate forces along all lateral axes. The limits of the lateral forces while hovering, with $\boldsymbol{\mu}_m = [0,0,0]^T$, are illustrated by the set $(P_{xy})$ of Fig. 3. These constraints change based on $f_{b_{z_{ref}}}$ computed from Eq. (21).

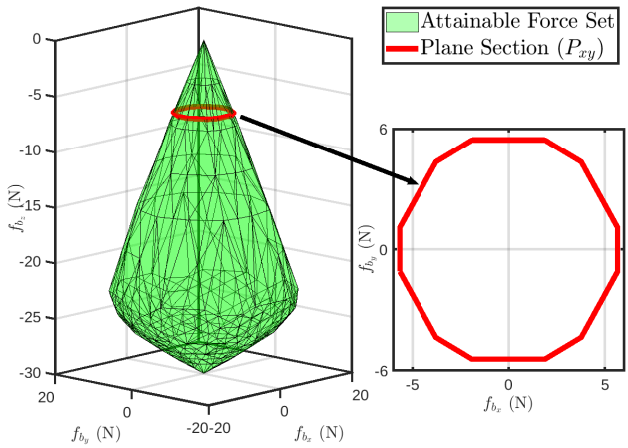

Fig. 3: The attainable force set of the simulated multirotor is shown in green, while the lateral force constraint plane ($P_{xy}$) during hovering with zero attitude is depicted in red.

To evaluate the physical limitations of the multirotor and its capability to perform different maneuvers, three main scenarios are considered in this work: 1) Hovering at different orientations, 2) Translating while maintaining minimal orientation, and 3) Tracking a full pose as described in Section II-C in the presence of disturbances.

### A. Scenario 1: Hovering with varying orientations

In this scenario, the objective is to hover at various orientations while maintaining accurate position tracking. Sinusoidal reference inputs are applied to the multirotor as $\boldsymbol{\mu}_{ref} = [0.7\sin(0.2t), -0.7\sin(0.5t), -0.35\sin(0.5t)]^T$ and $\boldsymbol{\xi}_{ref} = [0, 0, -3]^T$. The position tracking performance remains well-maintained, with the tracking error limited to 0.02 m, as shown in Fig. 4. Additionally, the attitude tracking demonstrates how $\boldsymbol{\mu}_{ref}$ is followed by both $\boldsymbol{\mu}_c$ and $\boldsymbol{\mu}_m$. It is observed that $\boldsymbol{\mu}_{ref} = \boldsymbol{\mu}_c$ whenever feasible. However, when $\boldsymbol{\mu}_{ref}$ is not achievable, an optimized feasible attitude is computed and tracked, as shown in Fig. 4 for $20\text{s} \leq t \leq 30\text{s}$. Meanwhile, $\boldsymbol{\mu}_m$ maintains accurate tracking of $\boldsymbol{\mu}_c$. The servo angles $\alpha_i$ and $\beta_i$ reach their saturation limits of 40° for $20\text{s} \leq t \leq 30\text{s}$ and within other time intervals when $\boldsymbol{\mu}_{ref}$ is infeasible, as shown in Fig. 5. This saturation results in constraints on the lateral forces, as illustrated in Fig. 6. Additionally, the 3D plot in Fig. 6 shows that the reference forces exceed the attainable force set, further confirming the saturation effects.

### B. Scenario 2: Translating with minimal orientation

In this scenario, the multirotor is required to track a sinusoidal reference position $\boldsymbol{\xi}_{ref} = [3\sin(\omega_x t), 2\sin(\omega_y t), -3]^T$ where the reference frequencies vary within different ranges $0 \text{ rad/s} \leq \omega_x \leq 0.5 \text{ rad/s}$, $0 \text{ rad/s} \leq \omega_y \leq 0.7 \text{ rad/s}$. The Euler angle references are kept at zero throughout the entire flight simulation with $\boldsymbol{\mu}_{ref} = [0, 0, 0]^T$. Figure 7 shows the tracking states of the multirotor where the plot of its position tracking shows good tracking. Additionally, it is shown that the attitude error is limited to $\pm 0.6°$. The variation between $\boldsymbol{\mu}_c$ and $\boldsymbol{\mu}_{ref}$ is mainly due to the weighted templates chosen for solving the guidance control allocation problem. In fact, since we are dealing with an optimization problem the optimal solution always depends on the chosen weights. Thus, as long as we penalize the second term of equation (17), we ensure minimizing the Euler angles. However, it is worth noting that over-penalizing the second term of equation (17) may cause the solution to relax the first term, which affects the performance of position tracking.

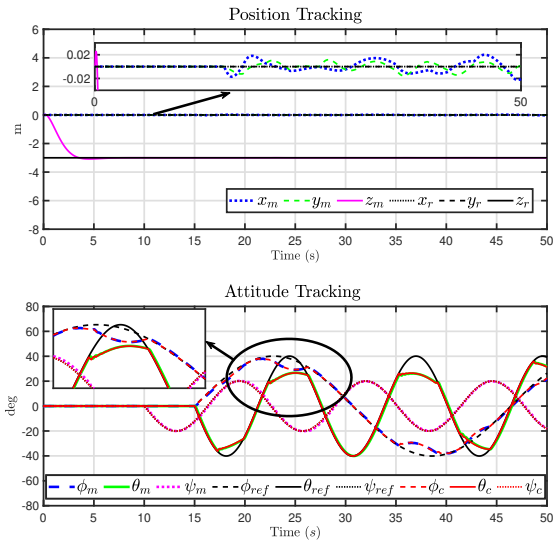

Fig. 4: Multirotor full-pose tracking while hovering with varying orientations.

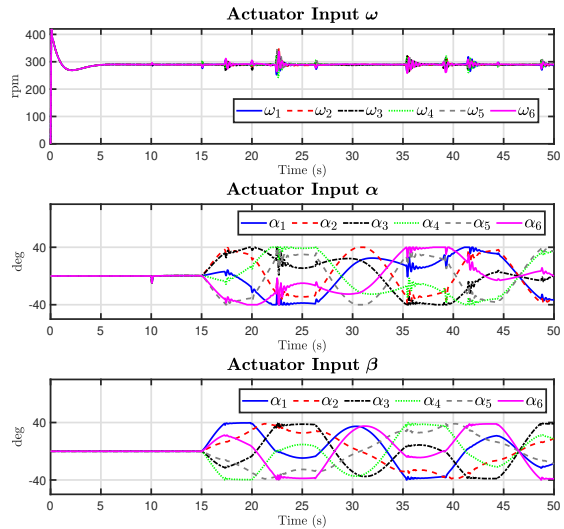

Fig. 5: Actuator control inputs of the multirotor while hovering with varying orientations.

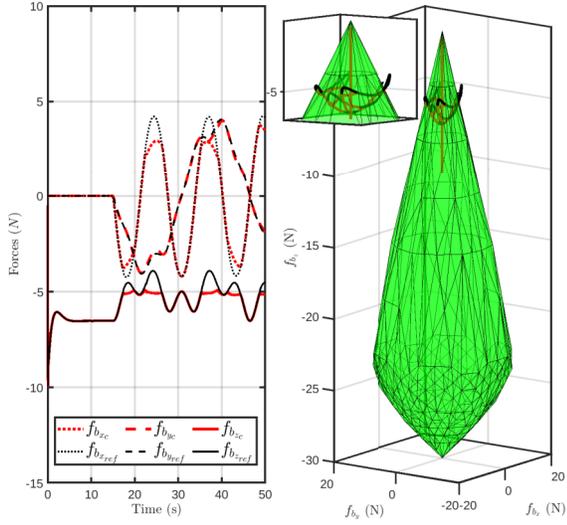

Fig. 6: Comparison between the reference forces $\boldsymbol{f}_{b_{ref}}$ and the commanded forces $\boldsymbol{f}_{b_c}$ while hovering at varying orientations.

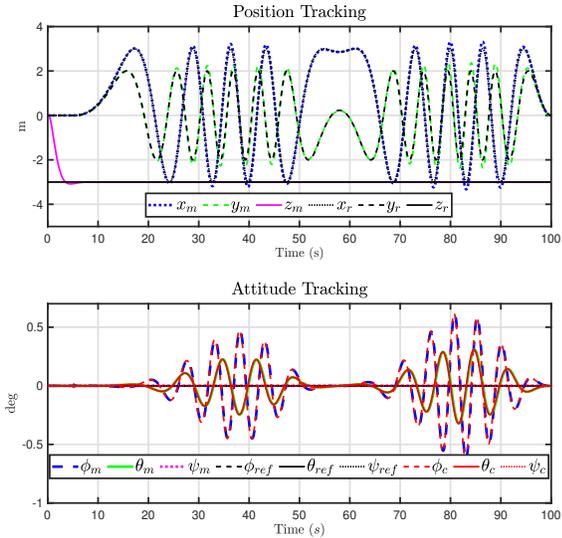

Fig. 7: Tracking performance of the multirotor for full-pose trajectory having minimal orientation.

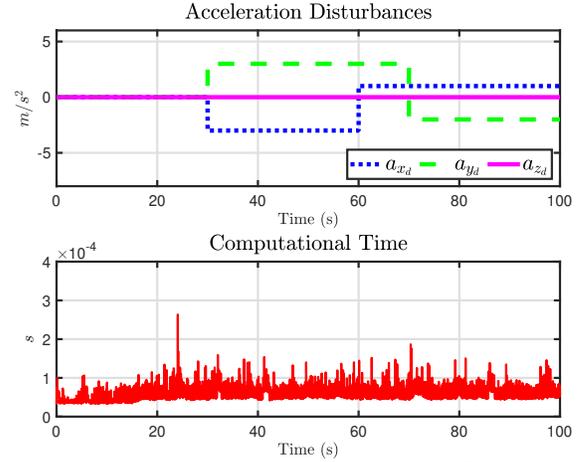

Fig. 8: The upper plot shows the disturbance on the multirotor's accelerations, while the lower shows the computation time for solving the guidance control allocation problem.

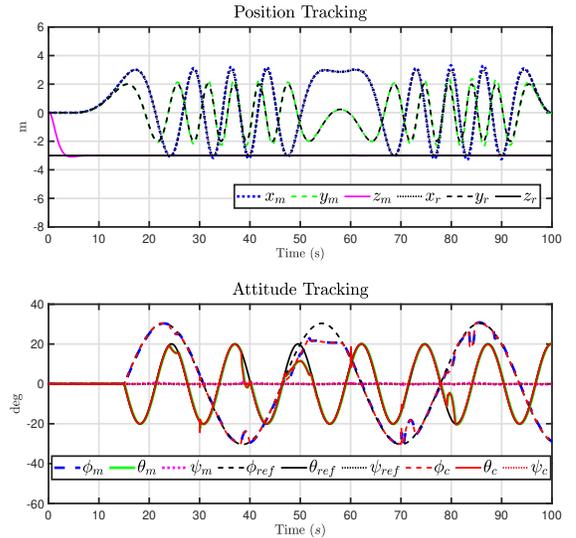

Fig. 9: Tracking performance of the multirotor for full-pose trajectory in the presence of disturbances.

### C. Scenario 3: Full-pose tracking with the presence of disturbances

In this scenario, the multirotor is asked to follow a position reference similar to Scenario 2. However, unlike before, the attitude reference now varies as $\boldsymbol{\mu}_{ref} = [0.53\sin(0.2t), -0.35\sin(0.5t), 0]^T$. Additionally, a disturbance signal in the form of a varying step function, shown in the upper plot of Fig. 8, is applied directly to the multirotor's acceleration output. It is introduced to account for external disturbances, which generate $\boldsymbol{f}_d$ that influence the multirotor's acceleration $\ddot{\boldsymbol{\xi}}_m$. Figure 9 presents the multirotor's full-pose tracking performance, demonstrating accurate position tracking even in the presence of disturbances. Whenever a disturbance occurs, its effect is primarily mitigated through adjustments in the attitude dynamics, as shown in the attitude tracking plot of Fig. 9. This behavior results from the selected guidance control allocation weights, emphasizing the importance of proper weight tuning to ensure good performance. On the other hand, the computational time required to solve the guidance control allocation optimization problem is shown in the lower plot of Fig. 8, with a maximum value of $2.5 \times 10^{-4}$ s. The actuator inputs to the system are presented in Fig. 10, where the servo angles reach saturation at different time intervals. Saturation primarily occurs when the multirotor operates near its control limits or in response to disturbances, as observed at $t = 70$ s.

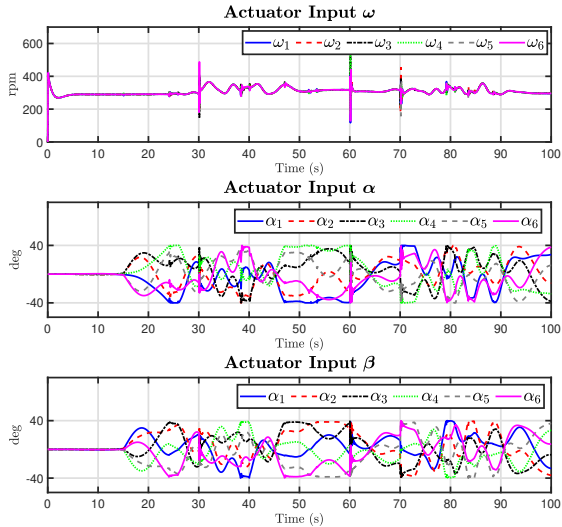

Fig. 10: Actuator control inputs of the multirotor during full-pose tracking with presence of disturbances.

## V. Conclusion and Future Work

The work proposes a cascaded control architecture for full-pose tracking of over-actuated multirotors. Robust controllers based on Incremental Nonlinear Dynamic Inversion (INDI) and structured $\mathcal{H}_\infty$ control are used to control both translational and rotational dynamics. Additionally, full-pose tracking (guidance loop) is formulated as a control allocation problem, and it is solved using an optimization-based weighted least-squares method. The proposed approach demonstrates effective performance across various tracking scenarios, including cases with external disturbances, contributing to the literature by enhancing robust control strategies for a wide range of multirotors specifically fully-actuated and over-actuated multirotors.

Future work will focus on the experimental validation of the proposed control method on an over-actuated multirotor. Ongoing efforts aim to implement the approach on an onboard flight controller using the Paparazzi open-source autopilot [22]. Further investigation is also required to evaluate the method's effectiveness in handling motor failures and recomputing the attainable force set, which is essential for limiting commanded lateral forces. Moreover, improvements to the inner control allocation strategy will be explored to ensure feasible actuator solutions and prevent saturation, particularly in over-actuated multirotors.

## References

[1] F. Ruggiero, V. Lippiello, and A. Ollero, "Aerial Manipulation: A Literature Review," IEEE Robotics and Automation Letters, vol. 3, no. 3, pp. 1957-1964, 2018, doi: 10.1109/LRA.2018.2808541.

[2] M. Brunner, G. Rizzi, M. Studiger, R. Siegwart, and M. Tognon, "A Planning-and-Control Framework for Aerial Manipulation of Articulated Objects," IEEE Robotics and Automation Letters, vol. 7, no. 4, pp. 10689-10696, 2022, doi: 10.1109/LRA.2022.3191178.

[3] M. Hamandi, F. Usai, Q. Sablé, N. Staub, M. Tognon, and A. Franchi, "Design of multirotor aerial vehicles: A taxonomy based on input allocation," The International Journal of Robotics Research, vol. 40, no. 8-9, pp. 1015-1044, 2021, doi: 10.1177/02783649211025998.

[4] A. Franchi, R. Carli, D. Bicego, and M. Ryll, "Full-Pose Tracking Control for Aerial Robotic Systems With Laterally Bounded Input Force," IEEE Transactions on Robotics, vol. 34, no. 2, pp. 534-541, 2018, doi: 10.1109/TRO.2017.2786734.

[5] M. Hamandi et al., "Full-Pose Trajectory Tracking of Overactuated Multi-Rotor Aerial Vehicles With Limited Actuation Abilities," IEEE Robotics and Automation Letters, vol. 8, no. 8, pp. 4951-4958, 2023, doi: 10.1109/LRA.2023.3290422.

[6] B. Convens, K. Merckaert, M. Nicotra, R. Naldi, and E. Garone, "Control of Fully Actuated Unmanned Aerial Vehicles with Actuator Saturation," IFAC-PapersOnLine, vol. 50, no. 1, pp. 12715-12720, 2017, doi: https://doi.org/10.1016/j.ifacol.2017.08.1823.

[7] M. Ryll, H. Bülthoff, and P. Giordano, "A Novel Overactuated Quadrotor Unmanned Aerial Vehicle: Modeling, Control, and Experimental Validation," IEEE Transactions on Control Systems Technology, vol. 23, no. 2, pp. 540-556, 2015, doi: 10.1109/TCST.2014.2330999.

[8] S. Rajappa, M. Ryll, H. H. Bülthoff and A. Franchi, "Modeling, control and design optimization for a fully-actuated hexarotor aerial vehicle with tilted propellers," Proceedings of the IEEE International Conference on Robotics and Automation, Seattle, WA, USA, pp. 4006-4013, 2015, doi: 10.1109/ICRA.2015.7139759.

[9] D. Invernizzi, M. Giurato, P. Gattazzo, and M. Lovera, "Comparison of Control Methods for Trajectory Tracking in Fully Actuated Unmanned Aerial Vehicles," IEEE Transactions on Control Systems Technology, vol. 29, no. 3, pp. 1147-1160, 2021, doi: 10.1109/TCST.2020.2992389.

[10] M. Hachem, C. Roos, and T. Miquel, "Application of Reduced-Order Robust Control to Multi-Rotor Stabilization and Guidance," Proceedings of the CEAS EuroGNC Conference, Bristol, UK, 2024, CEAS-GNC-2024-055.

[11] E. Smeur, G. Croon, and Q.P. Chu, "Cascaded Incremental Nonlinear Dynamic Inversion Control for MAV Disturbance Rejection," Control Engineering Practice, vol. 73, pp. 79-90, 2018, doi: 10.1016/j.conengprac.2018.01.003.

[12] M. Hachem, C. Roos, T. Miquel, and M. Bronz, "Improving Incremental Nonlinear Dynamic Inversion Robustness Using Robust Control in Aerial Robotics," arXiv preprint, 2025, arXiv:2501.07223.

[13] T. Johansen and T. Fossen, "Control allocation - A survey," Automatica, vol. 49, no. 5, pp. 1087-1103, 2013, doi: 10.1016/j.automatica.2013.01.035.

[14] M. Oppenheimer, D. Doman, and M. Bolender, "Chapter 8: Control allocation," in The Control Handbook, Control System Applications (Second Edition), W. Levine, Ed., CRC Press, 2010.

[15] Y. Su, P. Yu, M. Gerber, L. Ruan, and T.-C. Tsao, "Nullspace-Based Control Allocation of Overactuated UAV Platforms," IEEE Robotics and Automation Letters, vol. 6, no. 4, pp. 8094-8101, 2021, doi: 10.1109/LRA.2021.3103637.

[16] Y. Su, P. Yu, M. J. Gerber, L. Ruan, and T.-C. Tsao, "Fault-Tolerant Control of an Overactuated UAV Platform Built on Quadcopters and Passive Hinges," IEEE/ASME Transactions on Mechatronics, vol. 29, no. 1, pp. 602-613, 2024, doi: 10.1109/TMECH.2023.3288032.

[17] E. Sadien, C. Roos, A. Birouche, M. Carton, C. Grimault, L. Romana, and M. Basset, ""A detailed comparison of control allocation techniques on a realistic on-ground aircraft benchmark," Proceedings of the American Control Conference, Philadelphia, PA, USA, pp. 2891-2896, 2019, doi: 10.23919/ACC.2019.8814718.

[18] O. Härkegård, "Efficient active set algorithms for solving constrained least squares problems in aircraft control allocation," Proceedings of the 41st IEEE Conference on Decision and Control, Las Vegas, NV, USA, pp. 1295-1300 vol.2, 2002, doi: 10.1109/CDC.2002.1184694.

[19] E. Smeur, D. Höppener, and C. De Wagter, "Prioritized Control Allocation for Quadrotors Subject to Saturation,"

Proceedings of the International Micro Air Vehicle Conference, Toulouse, France, pp. 37–43, 2017, https://www.imavs.org/papers/2017/6.pd.
[20] A. Mancinelli, B. Remes, G. De Croon, and E. Smeur, "Real-Time Nonlinear Control Allocation Framework for Vehicles with Highly Nonlinear Effectors Subject to Saturation," Journal of Intelligent & Robotic Systems, vol. 108, no. 4, 2023, doi: 10.1007/s10846-023-01865-8.
[21] E. Baskaya, M. Hamandi, M. Bronz, and A. Franchi, "A Novel Robust Hexarotor Capable of Static Hovering in Presence of Propeller Failure," IEEE Robotics and Automation Letters, vol. 6, no. 2, pp. 4001-4008, 2021, doi: 10.1109/LRA.2021.3067182.
[22] G. Hattenberger, M. Bronz, and M. Gorraz, "Using the Paparazzi UAV System for Scientific Research," Proceedings of the International Micro Air Vehicle Conference and Competition, Delft, Netherlands, 2014, pp. 247-252.